\def\year{2017}\relax
\documentclass[letterpaper]{article}
\usepackage{aaai17}
\usepackage{times}
\usepackage{helvet}
\usepackage{courier}
\usepackage{graphicx}
\usepackage{amsmath}
\usepackage{multirow}
\usepackage{color}

\begin{document}
\title{Neural Models for Sequence Chunking}
\author{Feifei Zhai, Saloni Potdar, Bing Xiang, Bowen Zhou\\
IBM Watson \\
1101 Kitchawan Road, Yorktown Heights, NY 10598\\
\{fzhai,potdars,bingxia,zhou\}@us.ibm.com
}
\maketitle
\begin{abstract}
Many natural language understanding (NLU) tasks, such 
as shallow parsing (i.e., text chunking) and semantic slot 
filling, require the assignment of representative labels to the 
meaningful chunks in a sentence. Most of the current deep 
neural network (DNN) based methods consider these tasks
as a sequence labeling problem, in which a word, rather than a chunk, 
is treated as the basic unit for labeling. These chunks are then inferred 
by the standard IOB (Inside-Outside-Beginning) labels. 
In this paper, we propose an alternative approach 
by investigating the use of DNN for sequence chunking, 
and propose
three neural models 
so that each chunk can be treated as a complete unit for labeling.
Experimental results show that the proposed neural sequence chunking models
can achieve start-of-the-art performance on both the text chunking and 
slot filling tasks. 

\end{abstract}

\section{Introduction}

\begin{figure}
\begin{center}
\includegraphics[width=0.45\textwidth]{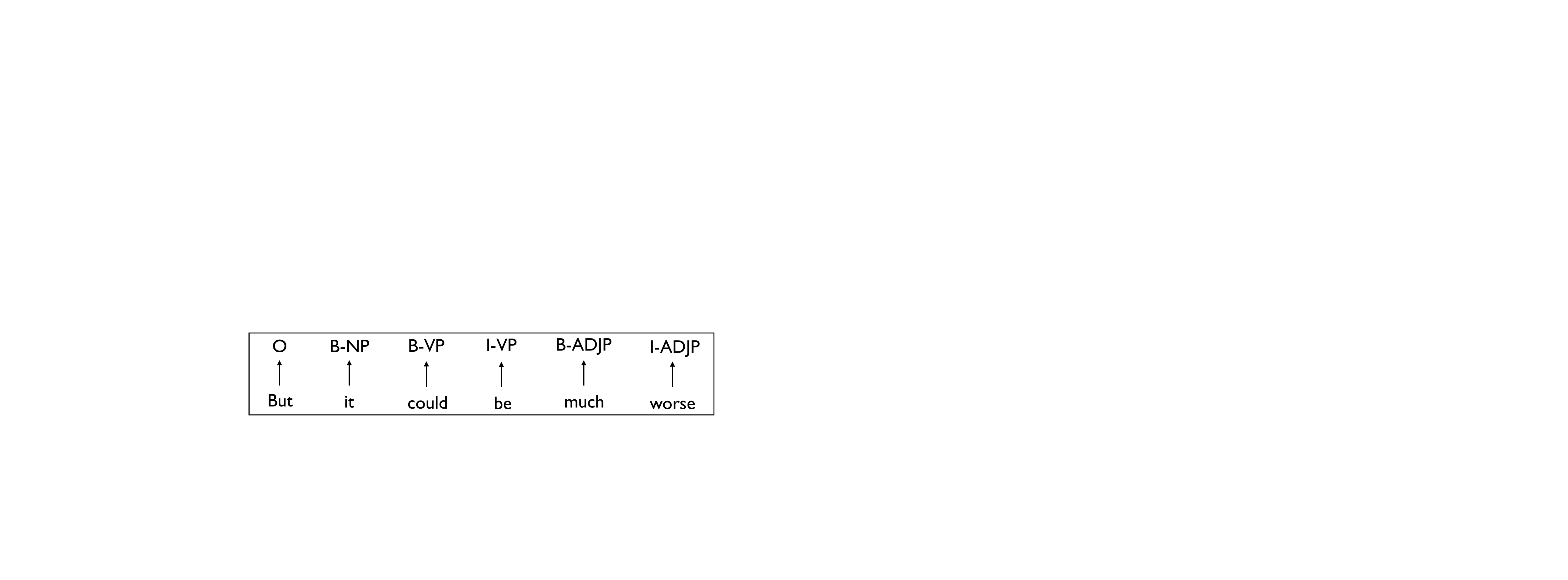}
\caption{
An example of text chunking where each word is labeled using the IOB scheme. The {\bf \textit{ chunk}} ``\textit{could be}'' is a verb phrase (VP) and ``\textit{it}'' is a {\bf \textit{single-word chunk}} with \textit{NP} label. 
\label{fig:example}}
\end{center}
\end{figure}

Semantic slot filling and shallow parsing which are standard NLU tasks fall under the umbrella of natural language understanding (NLU), which are usually solved by labeling meaningful chunks in a sentence.
This kind of task is usually treated as a sequence labeling problem, where every word in a sentence is assigned an IOB-based (Inside-Outside-Beginning) label.
For example, in Figure~~\ref{fig:example}, in the sentence ``\textit{But it could be much worse}'' we label ``\textit{could}'' as \textit{B-VP}, ``\textit{be}'' as \textit{I-VP}, and ``\textit{it}'' as \textit{B-NP}, while ``\textit{But}'' belongs to an artificial class \textit{O}.
This labeling indicates that a {\bf \textit{ chunk}} ``\textit{could be}'' is a verb phrase (VP) where the label prefix \textit{B} means the beginning word of the chunk, while \textit{I} refers to the other words within the same semantic chunk; and ``\textit{it}'' is a {\bf \textit{single-word chunk}} with \textit{NP} label. 

Such sequence labeling forms the basis for many recent deep network based approaches, e.g., convolutional neural networks (CNN), recurrent neural networks (RNN) or its variation, long short-term memory networks (LSTM). RNN and LSTM are good at capturing sequential information \cite{yao2013recurrent,huang2015bidirectional,mesnil2015using,peng2015recurrent,yang2016multi,kurata2016leveraging,zhu2016encoder}, 
whereas CNN can extract effective features for classification \cite{xu2013convolutional,vu2016sequential}.

Most of the current DNN based approaches use the IOB scheme to label chunks. However, this approach of these labels has a few drawbacks. First, we don't have an explicit model to learn 
and identify the scope of chunks in a sentence, instead we infer them implicitly (by IOB labels). 
Hence the learned model might not be able to fully utilize the training data which could result in poor performance. 
Second, some neural networks like RNN or LSTM have the ability to encode context information but don't treat each chunk as a complete unit. If we can eliminate this drawback, it could result in more accurate labeling, especially for multi-word chunks.

Sequence chunking is a natural solution to overcome the two drawbacks mentioned before. In sequence chunking, the original sequence labeling task is divided into two sub-tasks:
(1) {\bf \textit{Segmentation}}, to identify scope of the chunks explicitly;
(2) {\bf \textit{Labeling}}, to label each chunk as a single unit based on the segmentation results. 

\citeauthor{lample2016} (2016) used a stack-LSTM \cite{dyer2015} and a transition-based algorithm for sequence chunking. In their paper, the segmentation step is based on shift-reduce parser based actions. 
In this paper, we propose an alternative approach by relying only on the neural architectures for NLU. 
We investigate two different ways for segmentation: (1) using IOB labels; and (2) using pointer networks \cite{vinyals2015pointer} and propose three neural sequence chunking models. 
Pointer network performs better than the model using IOB. In addition, it also achieves state-of-the-art performance on both text chunking and slot filling tasks.

\section{Basic Neural Networks}
\label{sec:basic}

\subsection{Recurrent Neural Network}
Recurrent neural network (RNN) is a neural network that is suitable for modeling sequential information.
Although theoretically it is able to capture long-distance dependencies, in practice it suffers from the gradient vanishing/exploding problems \cite{bengio1994learning}.
Long short-term memory networks (LSTM) were introduced to cope with these gradient problems and model long-range dependencies \cite{hochreiter1997long} by using a memory cell. 
Given an input sentence $x = (x_1, x_2, ..., x_T)$ where $T$ is the sentence length,
LSTM hidden state at timestep $t$ is computed by: 
\begin{equation}
\label{eq:lstm}
\begin{split}
i_t &= \sigma(W^ix_t+U^ih_{t-1}+b^i) \\
f_t &= \sigma(W^fx_t+U^fh_{t-1}+b^f) \\
o_t &= \sigma(W^ox_t+U^oh_{t-1}+b^o) \\
g_t &= tanh(W^gx_t+U^gh_{t-1}+b^g) \\
c_t &= f_t\odot c_{t-1} + i_t \odot g_t\\
h_t &= o_t \odot tanh(c_t)
\end{split}
\end{equation}
where $\sigma(\cdot)$ and tanh$(\cdot)$ are the element-wise sigmoid and hyperbolic tangent functions, $\odot$ is the element-wise multiplication operator, and $i_t$,$f_t$,$o_t$ are 
the \textit{input}, \textit{forget} and \textit{output} gates.
$h_{t-1}$ and $c_{t-1}$ are the hidden state and memory cell of previous timestep respectively. To simplify the notation, we use $x_t$ to denote both the word and its embedding.

The bi-directional LSTM (Bi-LSTM), a modification of the LSTM, consists of a forward and a backward LSTM.
The forward LSTM reads the input sentence as it is (from $x_1$ to $x_T$) and computes the forward hidden states ($\overrightarrow{h_1}, \overrightarrow{h_2},..., \overrightarrow{h_T}$), while the backward LSTM reads the sentence in the reverse order (from $x_T$ to $x_1$), and creates backward hidden states ($\overleftarrow{h_1}, \overleftarrow{h_1},..., \overleftarrow{h_T}$).
Then for each timestep $t$, the hidden state of the Bi-LSTM is generated by concatenating $\overrightarrow{h_t}$ and $\overleftarrow{h_t}$,
\begin{equation}
\label{eq:blstm}
\overleftrightarrow{h_t} = [\overrightarrow{h_t}; \overleftarrow{h_t}]
\end{equation}

\subsection{Convolutional Neural Network}
Convolutional Neural Networks (CNN) 
have been used to
extract features for sentence classification \cite{kim2014convolutional,ma2015dependency,santos2015}.
Given a sentence, a CNN with \textit{m} filters and a filter  size of \textit{n} extracts a \textit{m}-dimension feature vector
from every \textit{n}-gram phrase of the sentence.
A \textit{max-over-time} pooling (max-pooling) layer is applied over all extracted feature vectors to create the final indicative feature vector (\textit{m}-dimension) for the sentence. 

Following this approach, we use CNN and max-pooling layer to extract features from chunks.
For each identified chunk, we first apply CNN to the embedding of its words (irrespective of it being a single-word chunk or  chunk), and then use the max-pooling layer on top to get the chunk feature vector for labeling. We use CNNMax to denote the two layers hereafter.

\section{Proposed Models}
\label{sec:models}

In this section, we introduce the different neural models for sequence chunking and discuss the final learning objective.

\subsection{Model I}
\label{sec:model_one}

\begin{figure}
\begin{center}
\includegraphics[width=0.4\textwidth]{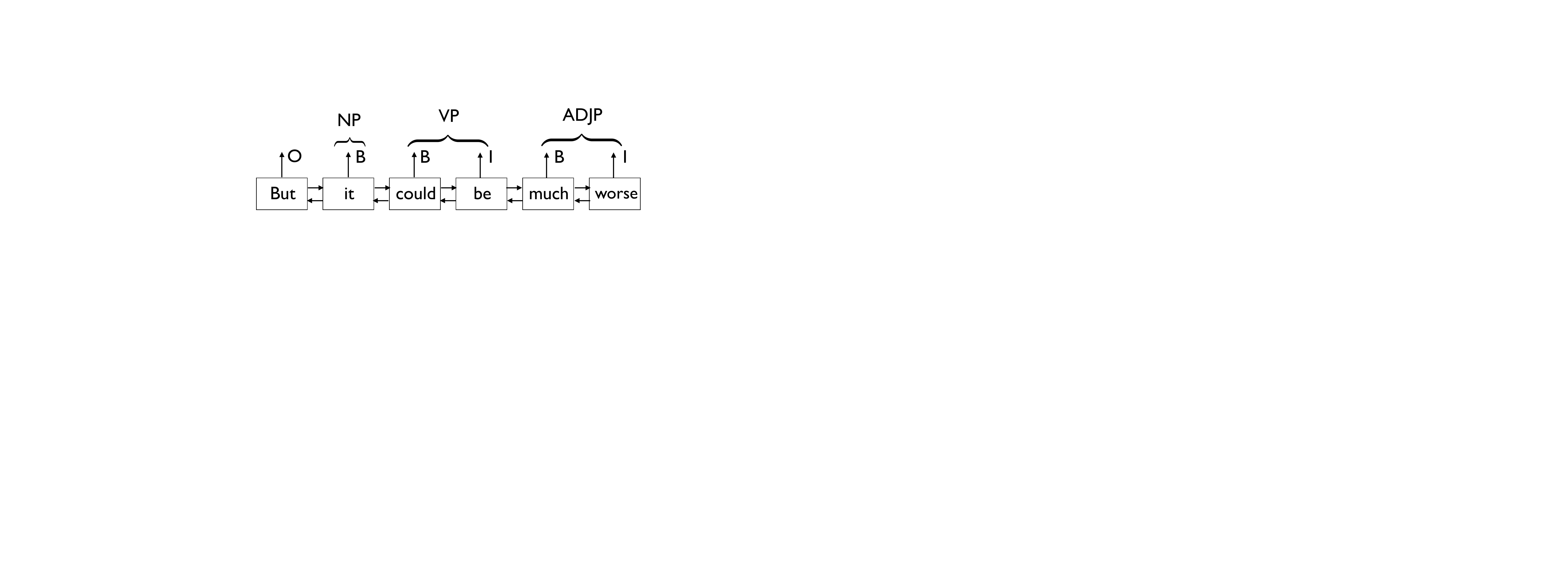}
\caption{Model I: Single Bi-LSTM model for both segmentation and labeling subtasks.
\label{fig:model_one}}
\end{center}
\end{figure}

For segmentation, the most straightforward and intuitive way is to transform it into a sequence labeling problem with 3 classes : I - inside, O - outside, B - beginning; and then understand the scope of the chunks from these labels. 
Building on this, we propose Model I, which is a Bi-LSTM as shown in Figure~\ref{fig:model_one}. 
In the model, we take the bi-LSTM hidden states generated by Formula~(\ref{eq:blstm}) as features for both segmentation and labeling. 

For example, we first classify each word into an IOB label as shown in (Figure~\ref{fig:model_one}).
Suppose a chunk begins at word $i$ with length $l$ (with one \textit{B} label and followed by $(l-1)$ \textit{I} labels), then we can compute a feature vector for a chunk as follows:

\begin{equation}
\label{eq:ave_hidden_chunk}
Ch_j = Average(\overleftrightarrow{h_i}, \overleftrightarrow{h_{i+1}} ,..., \overleftrightarrow{h_{i+l-1}})
\end{equation}

where $j$ is the chunk index of the sentence, and $Average(\cdot)$ computes the average of the input vectors. With $Ch_j$, we apply a softmax layer over all chunk labels for labeling.
For example in Figure~\ref{fig:model_one}, ``much worse'' is identified as a chunk with length 2; and we apply Formula~(\ref{eq:ave_hidden_chunk}) on its hidden states, to finally get the ``ADJP'' label.

\subsection{Model II}
\label{sec:model_two}

\begin{figure*}
\begin{center}
\includegraphics[width=0.75\textwidth]{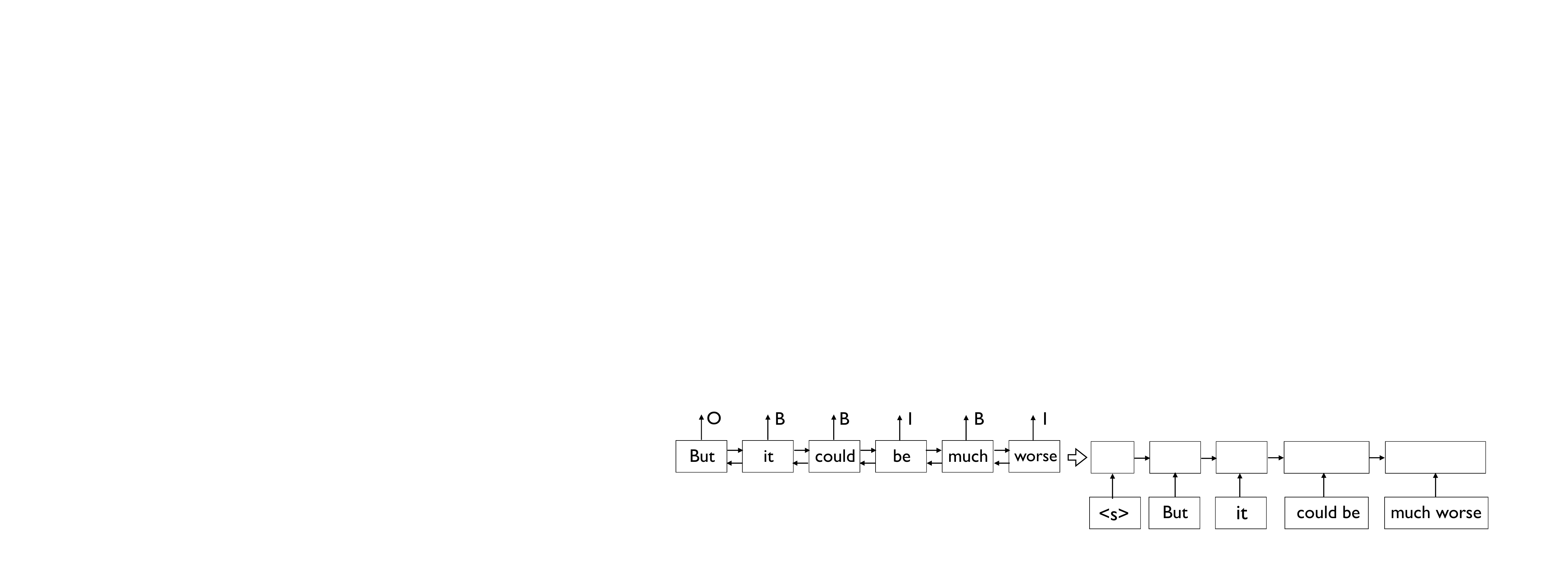}
\caption{
Model II: Encoder-decoder framework. The encoder Bi-LSTM is used for segmentation and the decoder LSTM is used for labeling.
\label{fig:model_two}}
\end{center}
\end{figure*}

A drawback of Model I is that a single Bi-LSTM may not perform well on both segmentation and labeling subtasks. 
To overcome this we propose Model II, which follows the encoder-decoder framework ( Figure~~\ref{fig:model_two}) \cite{sutskever2014sequence,bahdanau2014neural}.
Similar to Model I, we employ a Bi-LSTM for segmentation with IOB labels\footnote{Note that in Model I and II, 
we cannot guarantee that label ``\textit{O}'' 
is not followed by ``\textit{I}'' during segmentation. 
If so, we just take the first ``\textit{I}'' as ``\textit{B}''. In future work it is advisable to add that as a hard constraint.}.
This Bi-LSTM will also serve as an encoder and create a sentence representation 
$[\overrightarrow{h_T};\overleftarrow{h_1}]$
(by concatenating the final hidden state of the forward and backward LSTM) 
which is used to initialize the decoder LSTM.

We modify the general encoder-decoder framework and use chunks as the inputs instead of words. For example, \textit{much worse} is a  chunk 
in Figure~\ref{fig:model_two}, 
and we take it as a single input to the decoder. 
The input chunk representation $C_j$ consists of several parts.
We first use the CNNMax layer to 
extract important information from words inside the chunk:
\begin{equation}
\label{eq:cnnmax_word_chunk}
Cx_j = g(x_i, x_{i+1} ,..., x_{i+l-1})
\end{equation}
where $g(\cdot)$ is the CNNMax layer.
Then we use the context word embeddings of the chunk
to capture context information \cite{yao2013recurrent,mesnil2015using,kurata2016leveraging}.
The context window size is a hyperparameter to tune. 
Finally, we average the hidden states from the encoder Bi-LSTM by Formula~(\ref{eq:ave_hidden_chunk}).
By using these three parts, 
we extract different useful information for labeling,
and import them all
into the decoder LSTM. 
Thus, the decoder LSTM hidden state is updated by:
\begin{equation}
\label{eq:model_two_lstm}
h_j = \textit{LSTM}(Cx_j, Ch_j, Cw_j, h_{j-1}, c_{j-1})
\end{equation}
here $Cw_j$ is the concatenation of context word embeddings. 
Note that the computation of hidden states here is similar to Formula~(\ref{eq:lstm}),
the only difference is that here we have three inputs $\{Cx_j, Ch_j, Cw_j\}$. 
The generated hidden states are finally used for labeling 
by a softmax layer.

\begin{figure*}
\begin{center}
\includegraphics[width=0.75\textwidth]{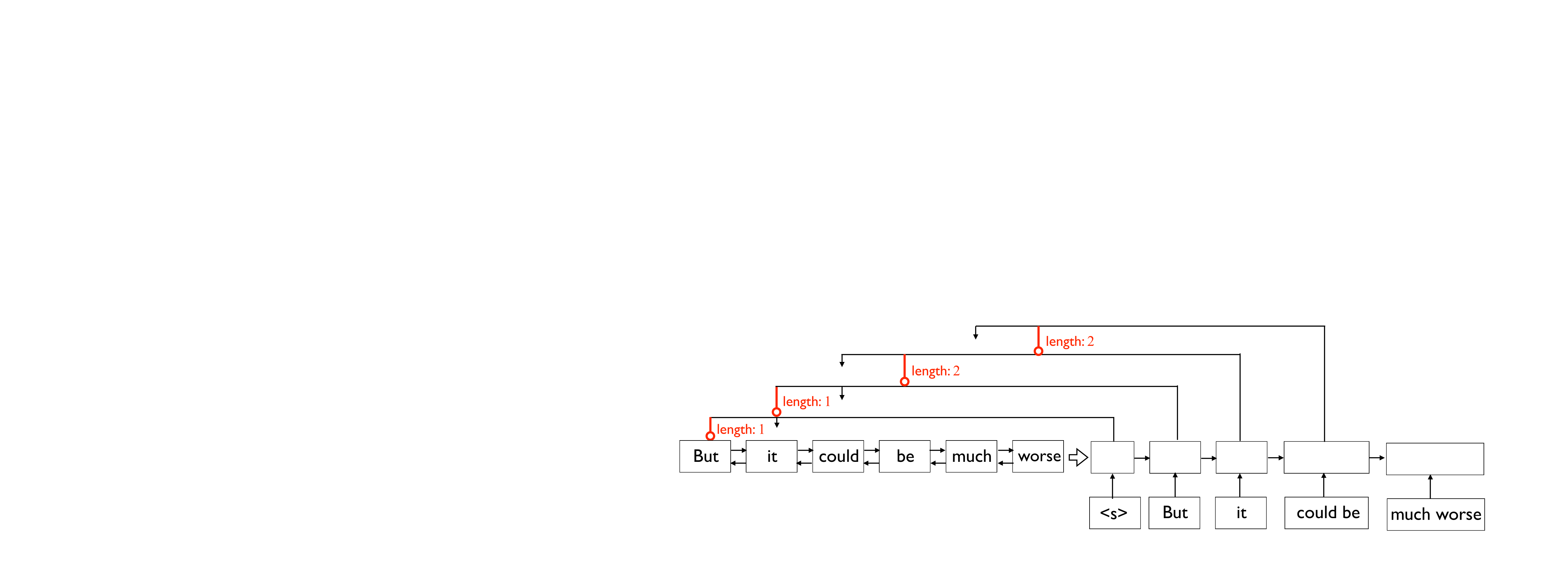}
\caption{
Model III: Encoder-decoder-pointer framework : Segmentation is done by a pointer network and a decoder LSTM is used for labeling.
\label{fig:model_three}}
\end{center}
\end{figure*}

\subsection{Model III}
\label{sec:model_three}

%The drawback of using IOB labels to identify the scope of chunks
%is that it is hard to use chunk-level features for segmentation, 
%like the length of chunks. 
%Similarly, 
%the shift-reduce algorithm used in \cite{lample2016}
%also cannot use chunk level features. 
There are two drawbacks of using IOB labels for segmentation.
First, 
it is hard to use chunk-level features for segmentation, 
like the length of chunks. 
Also,
using IOB labels cannot compare different chunks directly. 
The shift-reduce algorithm used in \cite{lample2016} has the same issue. 
They both transform a multi-class classification problem (we could have a lot of chunk candidates)
into a 3-class classification problem, in which the chunks are  inferred implicitly.

To resolve this problem, we further propose Model III, 
which is an encoder-decoder-pointer framework (Figure~\ref{fig:model_three}) \cite{nallapati:CoNLL}.
Model III is similar to Model II, the only difference being the method of identifying chunks.

Model III is a greedy process of segmentation and labeling, 
where we first identify one chunk,
and then label it. This process is repeated until all the words are processed.
As all chunks are adjacent to each other 
\footnote{
Here as we don't know the label of each chunk during segmentation, 
we need to feed all the chunks to the decoder for labeling.	
},
after one chunk is identified, 
the beginning point of the next one is also known,
and only its ending point is to be determined.
We adopt pointer network \cite{vinyals2015pointer}
to do this. 
For a possible chunk beginning at timestep $b$,
we first generate a feature vector 
for each possible ending point candidate $i$:
\begin{equation}
\begin{split}
  u^i_j =&\ v_1^Ttanh(W_1\overleftrightarrow{h_i} + W_2x_i + W_3x_b + W_4d_j) \\
        & +\ v_2^TLE(i-b+1)  \ \ \ \ \ \ \ \ \  \ \ \ \ \ \  i \in [b, b+l_m)
\end{split}
\end{equation}
where $j$ is the decoder timestep (i.e., chunk index), 
$l_m$ is the maximum chunk length.
We use the encoder hidden state $\overleftrightarrow{h_i}$,
the ending point candidate word embedding $x_i$,  
together with current beginning word embedding $x_b$
and decoder hidden state $d_j$ as features. 
We also use the chunk length embedding, $LE(i-b+1)$, 
as the chunk level feature.
$W_1$,$W_2$,$W_3$,$W_4$,$v_1$,$v_2$ and $LE$
are all learnable parameters. 
Then the probability of choosing ending point candidate $i$ is:
\begin{equation}
p(i) = \frac{exp(u^i_j)}{\sum_{k=b}^{b+l_m-1} exp(u^k_j)}
\end{equation}
We use this probability to identify the scope of chunks. 
For example, 
suppose we just identified word \textit{it} as a one word chunk with label \textit{NP} in Figure~\ref{fig:model_three}.
Following the line emitted from it, 
we will need to decide the ending point of the next chunk
(the beginning point is obviously the word \textit{could} after \textit{it}).
With the maximum chunk length 2, 
we have two choices, 
one is to stop at word \textit{could} and gets a one word chunk \textit{could},
and the other is to stop at word \textit{be}
and generates a two word chunk \textit{could be}.
From the figure, we can see that the model selects the second case
(red circle part),
and creates a two word chunk.
This chunk
will serve as the input of the next decoder timestep.
The decoder hidden states are updated similar to Model II (Equation~\ref{eq:model_two_lstm}).

\subsection{Learning Objective}
As we described above, all the aforementioned models solve two subtasks - segmentation and labeling. 
We use the cross-entropy loss function for both the two subtasks, and sum the two losses to form the the learning objective:
\begin{equation}
\label{eq:objective}
L(\theta) = L_{segmentation}(\theta) + L_{labeling}(\theta)
\end{equation}
where $\theta$ denotes the learnable parameters. Alternatively, we could also use weighted sum, or do multi-task learning by considering segmentation and labeling as the two tasks. We leave these extensions as future work.

\section{Experiments}
\label{sec:experiments}

\subsection{Experimental Setup}
\begin{figure}
\begin{center}
\includegraphics[width=0.45\textwidth]{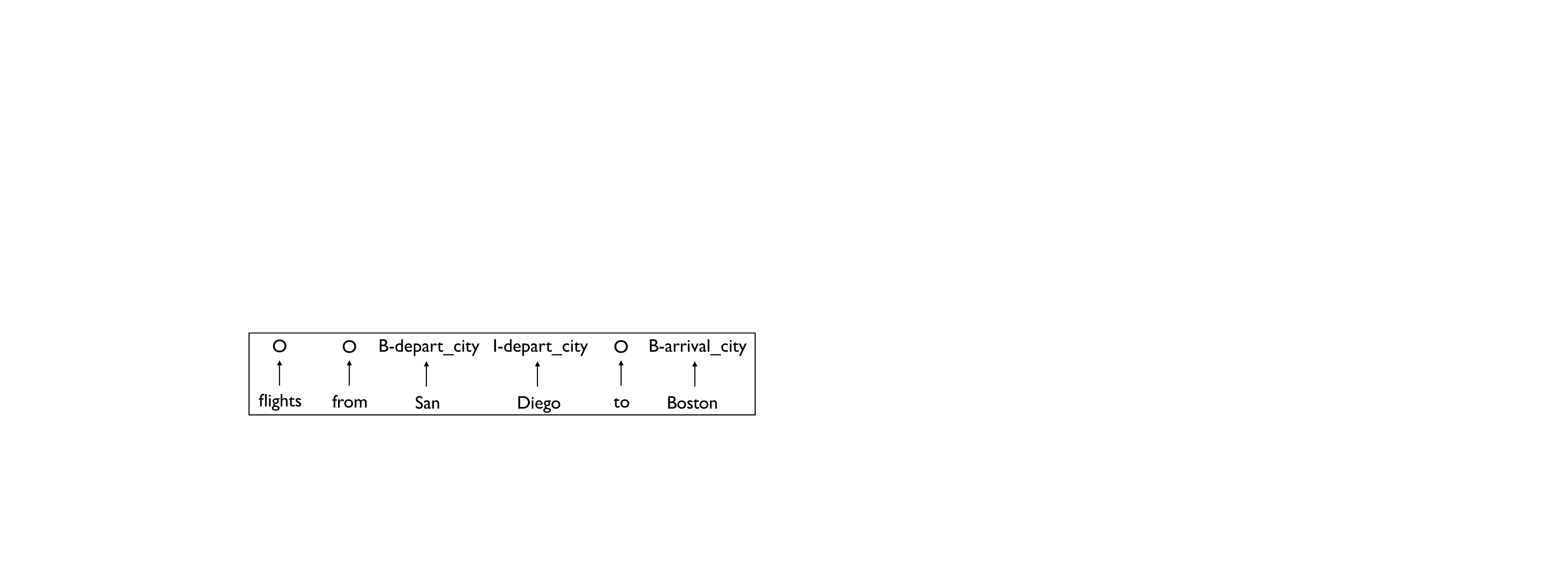}
\caption{
An example of semantic slot filling using the IOB scheme. \textit{``San Deigo''} is a \bf{\textit{multi-word chunk}} with label \textit{``depart\_city''}.
\label{fig:slot_example}}
\end{center}
\end{figure}

We conduct experiments on text chunking and semantic slot filling respectively
to test the performance of the neural sequence chunking models we propose in this paper.
Both these tasks identify the meaningful chunks in the sentence, 
such as the noun phrase (NP), or the verb phrase (VP) for text chunking in Figure~\ref{fig:example}, and the \textit{``depart\_city''} for slot filling task in Figure~\ref{fig:slot_example}.

We use the CoNLL 2000 shared task \cite{tjong2000introduction} dataset for text chunking.
It contains 8,936 training and 893 test sentences.
There are 12 different labels (22 with IOB prefix included).
Since it doesn't have a validation set, we hold out 10\% of the training data (selected at random) as the validation set.

To evaluate the effectiveness of our method on the semantic slot filling task, we use two different datasets. The first one is the ATIS dataset, which consists of reservation requests from the air travel domain. 
It contains 4,978 training and 893 testing sentences in total, with a vocabulary size of 572. 
There are 84 different slot labels (127 if with IOB prefix). 
We randomly selected 80\% of the training data for model training and the rest 20\% as the validation set \cite{mesnil2015using}.
Following the work of \cite{kurata2016leveraging}, we also use a larger dataset by combining the ATIS corpus with the MIT Restaurant Corpus and MIT Movie Corpus \cite{liu2013asgard,liu2013query}.
This dataset has 30,229 training and 6,810 testing instances. Similar to the previous dataset, we use 80\%  of the training instances for training the model, and treat the rest 20\% as a validation set.
This dataset has a vocabulary size of 16,049 and the number of slot labels is 116 (191 with IOB prefix included). Since this dataset is considerably larger and includes 3 different domains, we use ``LARGE'' to denote it hereafter.

The final performance is measured in terms of F1-score,
computed by the public available script \textit{conlleval.pl} 
\footnote{http://www.cnts.ua.ac.be/conll2000/chunking/}.
We report the F1-score on the test set with
parameters that achieves the best F1-score on the validation set.
Towards the neural sequence chunking models, 
after we get the label for each chunk,
we will assign each of its word an 
IOB-based label accordingly so that the 
script can do evaluation. 
We also report the segmentation F1-score
to assess the segmentation performance of different models.
This is also computed by the \textit{conlleval.pl} script,
but only considers three labels, i.e. \{I,O,B\}.
To compute the segmetnation F1-score, 
we delete the content label for each word,
for example, 
if a word has a label ``\textit{B-VP}'',
we will delete ``\textit{VP}''
and the left ``\textit{B}'' is used 
for segmentation F1-score.

For the two tasks, 
we use hidden state size as 100 
for the forward and backward LSTM respectively in Bi-LSTM,
and size 200 for the LSTM decoder. 
We use dropout with rate 0.5 on 
both the input and output of all LSTMs. 
The mini-batch size is set to 1. 
The number of training epochs are limited to 200 for text chunking,
and 100 for slot filling.
\footnote{We found that while 100 epochs are enough for slot filling model to converge, we need 200 for text chunking.}
For the CNN used in Model II and III on extracting chunk features,
the filter size is the same as word embedding dimension,
and the filter window size as 2.
We adopt SGD to train the model,
and by grid search, 
we tune the initial learning rate in [0.01, 0.1],
learning rate decay in [1e-6, 1e-4],
and context window size \{1,3,5\}.

For the word embedding,
following \cite{kurata2016leveraging},
we don't use pre-trained embedding for the slot filling task, 
but use a randomly initialized embedding 
and tune the dimension in $\{30, 50, 75\}$ by grid search.
For text chunking, 
we concatenate two different embeddings.
The first is %based on the result of \cite{huang2015bidirectional},
SENNA embedding \cite{collobert2011natural} 
with dimension 50. \footnote{http://ronan.collobert.com/senna/} 
The other is a word representation generated 
based on its composed characters.
we adopt a CNN onto the randomly initialized character embeddings,
with 30 filters and filter window size 3.

\subsection{Text Chunking Results}
Results on the text chunking task are shown in Table~\ref{table:np-chunking-result}. 
In this, the ``baseline (Bi-LSTM)'' refers to a Bi-LSTM model for sequence labeling 
(use IOB-based labels on words as in Figure~\ref{fig:example}).
``F1'' is the final evaluation metric, and ``segment-F1'' refers to the segmentation F1-score. 
From the table, we can see that Model I and Model II only have comparable results with 
the baseline on both evaluation metrics - segment-F1 and final F1 score.
Hence, we infer that using IOB labels to do segmentation independently might not be a good choice.
However, Model III outperforms the baseline on both segmentation and labeling. 

\begin{table}[ht!]
\begin{centering}
\begin{tabular}{c|c|c}
\hline
					& F1 & Segment-F1        \\
\hline %\hline
baseline (Bi-LSTM)  & 94.13 & 95.28 \\
Model I   			& 94.01 & 95.09 \\
Model II   			& 94.13 & 95.22 \\
Model III 			& \bf{94.72} & \bf{95.75}\\
\hline
\end{tabular}
\vspace{-0.1cm}
\caption{Text chunking results 
of our neural sequence chunking models. 
}
\label{table:np-chunking-result}
\end{centering}
\end{table}

We further compare our best result 
with the current published results 
in Table~\ref{table:textchunking-compare-published}.
In the table,
\cite{collobert2011natural}
is the first work of using neural networks for text chunking. \citeauthor{huang2015bidirectional} used a BiLSTM-CRF framework
together with a lot of handcraft features. 
\citeauthor{yang2016multi} extend this framework
and employ a GRU to incorporate the character information of words,
rather than using handcrafted features. 
To our best knowledge, 
they got the current best results 94.66 
on the text chunking task.
\footnote{They also get a performance of 95.41, 
but this number is from joint training, 
which needs the training data of other tasks.}
Different from previous work,
we model the segmentation part explicitly 
in our neural models,
and without using CRF, 
we get a state-of-the-art performance of 94.72.

\begin{table}[ht!]
\begin{centering}
\scalebox{0.9}{
\begin{tabular}{c|c}
\hline
Methods & F1-score \\
\hline \hline
SVM Classifier \cite{kudo2000}    &  93.48  \\
SVM Classifier \cite{kudo2001chunking}    &  93.91  \\
Second order CRF \cite{sha2003shallow} & 94.30 \\
HMM + voting scheme \cite{shen2005voting} & 94.01 \\
Conv network tagger (senna) \cite{collobert2011natural} & 94.32 \\
BiLSTM-CRF \cite{huang2015bidirectional} & 94.46  \\
BiGRU-CRF \cite{yang2016multi}\footnote & 94.66 \\
\hline
\hline
Model III (Ours) & \bf{94.72}\\
\hline
\end{tabular}}
\caption{
Comparison with published results on the CoNLL chunking dataset.
}
\label{table:textchunking-compare-published}
\end{centering}
\end{table}

\subsection{Slot Filling Results}
\subsubsection{Segmentation Results}

\begin{table}[ht!]
\begin{centering}
\scalebox{0.8}{
\begin{tabular}{c|c|c|c|c}
\hline
\multirow{2}{*}{} 	& \multicolumn{2}{c|}{ATIS}        & \multicolumn{2}{c}{LARGE} \\ 
\cline{2-5}
					& F1 & Segment-F1          & F1 & Segment-F1\\
\hline %\hline
baseline (Bi-LSTM)  & 95.23 & 98.85 & 75.73 & 80.79 \\
Model I   			& 95.25 & 98.92 & 76.68 & 79.93 \\
Model II   			& 95.71 & 98.82 & 77.26 & 79.99 \\
Model III 			& \bf{95.86} & \bf{99.01} & \bf{78.49} & \bf{82.44}\\
\hline
\end{tabular}}
\caption{Main results of our neural sequence chunking models on slot filling task. 
}
\label{table:main-result}
\end{centering}
\end{table}

From the Table~\ref{table:main-result}, we can see that 
the segment-F1 score on ATIS data
is much better than the one on LARGE data
($\sim$99\% vs. $\sim$80\%).
This is because the ATIS data is much easier 
for segmentation than LARGE data.
As shown in Table~\ref{table:chunk-len-statistic},
more than 97\% of the chunks in ATIS data
have only one or two words, 
while the LARGE data has much longer chunks. 
Also, compared to the small ATIS vocabulary (572 words),
it is harder to learn 
a good segmentation model with 
a more complicated vocabulary (about 16k words)
in LARGE data.
%lists the statistics of chunks 
%with different length in ATIS data.
%with only 572 words vocabulary,
%For short chunks, 
%it would be very easy to identify them
%while for longer ones, 
%it should be harder and thus get bad results.
%our sequence chunking model (model III) performs 
%much better (table~\ref{table:chunk-len-Segment-f1}).

\begin{table}[ht!]
\begin{centering}
\scalebox{0.75}{
\begin{tabular}{c|c|c|c|c}
\hline
\multirow{2}{*}{} 	& \multicolumn{2}{c|}{ATIS}        & \multicolumn{2}{c}{LARGE} \\ 
\cline{2-5}
					& Train & Test          & Train & Test\\
\hline
1 & 10275 (77.7\%) & 2096 (73.9\%)& 28511 (46.8\%) & 7283 (42.8\%)\\
\hline
2 & 2726 (20.6\%) & 659 (23.2\%)& 20679 (34.0\%) & 6214 (36.5\%)\\
\hline
$>=$3 & 224 (1.7\%) & 82 (2.9\%)& 11694 (19.2\%) & 3516 (20.7\%) \\
\hline
\end{tabular}}
\caption{
Statistics on the length of chunks: The first column denotes chunk-lengths. 
For example, first cell indicates that 
there are 10275 chunks of length 1, 
and accounts for 77.7\% of all ATIS chunks. 
}
\label{table:chunk-len-statistic}
\end{centering}
\end{table}

Moreover, Model III gets the best segmentation performance over all the models
(99.01\% and 82.44\%),
confirming that our pointer network in model III is good at
this task.
However, Model I and II are comparable to baseline 
on the easy ATIS data, and are about 1\% worse on LARGE data.
This further confirms our analysis on text chunking experiments 
that 
using IOB labels alone for segmentation, 
(like in Model I and II)
cannot give us a good result. 

\begin{table}[ht!]
\begin{centering}
\scalebox{0.75}{
\begin{tabular}{c|c|c|c|c|c|c}
\hline
\multirow{2}{*}{} 	& \multicolumn{3}{c|}{ATIS}        & \multicolumn{3}{c}{LARGE} \\ 
\cline{2-7}
					& 1 & 2 &$>=$3 & 1 & 2 & $>=$3 \\
\hline
Baseline(Bi-LSTM) & 98.90 & 98.70 & 98.78 & 86.25 & 88.48 & 54.88\\
\hline
Model I & 98.95 & 98.78 & 99.39 & 85.91 & 87.27 & 53.30\\
\hline
Model II & 98.83 & 98.78 & 98.78 & 86.42 & 87.29 & 52.98\\ 
\hline
Model III & \bf{99.00} & \bf{98.93} & \bf{100.0} & \bf{89.01} & \bf{88.69} & \bf{56.59}\\
\hline
\end{tabular}}
\caption{
Segment-F1 on different chunk-lengths. 
}
\label{table:chunk-len-segment-f1}
\end{centering}
\end{table}
We further investigate the segmentation process and show the segmentation F1-score on different chunk lengths in Table~\ref{table:chunk-len-segment-f1}.
The results demonstrate that 
the poor performance on LARGE data is mainly due to
the bad performance on identifying long chunks (around 55\%).
Our Model III improves this score by
~2\% over baseline (54.88\% vs. 56.59\%). As the absolute performance on this subset is still low, future research efforts should focus on improving this performance.
In addition, 
Model I and II get comparable segmentation results 
with the baseline model on one-words chunks,
while being worse on longer chunks,
further supporting this analysis.

\subsubsection{Labeling Results}

From Table~\ref{table:main-result},
we observe that 
Model III has the best F1 score as compared to the baseline and other neural chunking models. 
Another observation is that Model I and II  get better improvements over baseline even though they are poor at segmentation in slot filling task.

\begin{table}[ht!]
\centering
\scalebox{0.75}{
\begin{tabular}{c|c|c|c|c|c|c}
\hline
\multirow{2}{*}{} 	& \multicolumn{3}{c|}{ATIS}        & \multicolumn{3}{c}{LARGE} \\ 
\cline{2-7}
					& 1 & 2 &$>=$3 & 1 & 2 & $>=$3 \\
\hline
baseline(Bi-LSTM) & 95.37 & 96.03 & 85.19 & 79.21 & 83.56 & 53.36\\
\hline
Model I & 95.23 & 96.18 & 88.48 & 83.10 & 82.35 & 51.76\\
\hline
Model II & 95.87 & 96.18 & 87.80 & 85.01 & 82.82 & 51.15\\ 
\hline
Model III & \bf{95.89} & \bf{96.19} & \bf{92.68} & \bf{84.97} & \bf{83.89} & \bf{54.38}\\
\hline
\end{tabular}}
\caption{
F1-scores for different chunk-lengths
}
\label{table:chunk-len-f1}
\end{table}

Table~\ref{table:chunk-len-f1} gives some insights on this
by showing the F1-score on different chunk-lengths.
Comparing Table~\ref{table:chunk-len-segment-f1} and \ref{table:chunk-len-f1},
we can see when Model I and II achieve comparable segment-F1 with baseline, and the F-1 scores are higher.
%(all chunks in ATIS data and the one-word chunks in LARGE data). 
%The improvement of Model I and II over baseline should come from sequence chunking.
%We think this is because 
For slot filling task, 
the joint learning framework (Formula~(\ref{eq:objective}))
helps labeling while harms segmentation on model I and II.
Moreover, 
the usage of encoder in Model II could also help labeling
in this task
\cite{kurata2016leveraging}.
Finally, our Model III could achieve better F1 score on all chunk lengths.

\subsubsection{Comparison with Published Results}

We compare the ATIS results of our best model (Model III) 
with current published results in Table~\ref{table:atis-compare-published}.
As shown in the table, many researchers have done a lot of work which uses deep neural networks for slot filling.
Recent work shows the ranking loss is helpful \cite{vu2016bi}, and adding encoder improves the score to 95.66\%. The best published result in the table is from \cite{zhu2016encoder}, which is 95.79\%. Compared with previous results, 
our Model III gets the state-of-the-art performance 95.86\%.

\begin{table}[ht!]
\begin{centering}
\scalebox{0.9}{
\begin{tabular}{c|c}
\hline
Methods & F1-score \\
\hline \hline
RNN \cite{yao2013recurrent}    &  94.11  \\
CNN-CRF \cite{xu2013convolutional} & 94.35   \\
Bi-RNN \cite{mesnil2015using}  & 94.73  \\
LSTM \cite{yao2014spoken}  & 94.85 \\
RNN-SOP \cite{liu2015recurrent} & 94.89  \\
Deep LSTM \cite{yao2014spoken} & 95.08  \\
RNN-EM \cite{peng2015recurrent} &  95.25 \\
Bi-RNN with ranking loss \cite{vu2016bi} & 95.56  \\
Sequential CNN \cite{vu2016sequential} & 95.61  \\
Encoder-labeler Deep LSTM \cite{kurata2016leveraging} & 95.66  \\
BiLSTM-LSTM (focus) \cite{zhu2016encoder} & 95.79  \\
\hline
\hline
Model III (Ours) & \bf{95.86}\\
\hline
\end{tabular}}
\caption{
Comparison with published results on the ATIS data
}
\label{table:atis-compare-published}
\end{centering}
\end{table}

We compare our approach against the only set of published results on the LARGE data from \cite{kurata2016leveraging}, against which we compare our approach. 
The reported F1 score on this dataset by their encoder-decoder model is 74.41, and our best model achieves a score of 78.49 which is significantly higher.

\section{Related Work}

%Text chunking has attracted lots of interests in the machine learning and natural language processing community. Many methods such as Conditional Random Fields (CRFs), Hidden Markov Models (HMMs) and ensembling, have been explored to solve this problem \cite{kudo2001chunking,sha2003shallow,mcdonald2005flexible,shen2005voting,sun2008modeling}. 

In recent years, 
many deep learning approaches have been explored for resolving the sequence labeling tasks. 
\cite{collobert2011natural} proposed an effective window-based approach, in which they used 
a feed-forward neural network to classify each word and conditional random fields (CRF) to capture the sequential information. 
CNNs are also widely used for extracting effective classification features \cite{xu2013convolutional,vu2016sequential}. 

RNNs are a straightforward and better suited choice for these tasks as they model sequential information.
\cite{huang2015bidirectional} presented a BiLSTM-CRF model, 
and achieved state-of-the-art performance on several tasks, 
like named entity recognition and text chunking with the help of handcrafted features. 
\cite{chiu2015named} used a BiLSTM for labeling and a CNN to capture character-level information,
like \cite{dos2014deep} and additionally used handcrafted features to gain good performance. 
Many works have then been investigated to combine the advantages of the above two works 
and achieved state-of-the-art performance without handcrafted features.
These works usually use a BiLSTM or BiGRU as the major labeling architecture,
and a LSTM or GRU or CNN to capture the character-level information,  
and finally a CRF layer to model the label dependency \cite{lample2016,ma2016end,yang2016multi}. 

In addition, many similar works have also been explored for slot filling, %by using RNN-based networks,
like RNN \cite{yao2013recurrent,mesnil2015using}, LSTM \cite{yao2014spoken,jaech2016domain},
adding external memory \cite{peng2015recurrent}, adding encoder \cite{kurata2016leveraging},
using ranking loss \cite{vu2016bi}, adding attention \cite{zhu2016encoder} and so on.

In the other direction, people also developed neural networks 
to help traditional sequence processing methods, 
like CRF parsing \cite{durrett2015} and weighted finite-state transducer \cite{rastogi2016}.

%Our work differs from previous works by dividing the original task into two subtasks and 
%model them inside one neural architecture. 

%There are also many works on slot filling by using RNN or some variations to incorporate different useful information
%\cite{yao2013recurrent,yao2014spoken,mesnil2015using,peng2015recurrent,kurata2016leveraging,jaech2016domain,zhu2016encoder,vukotic2016step}.
%Our work differs from previous work by 
%First we model the segmentation step explicitly by neural architectures.

\section{Conclusion}
\label{sec:copyright}
In this paper, we presented three different models for sequence chunking.
Our experiments show that the segmentation results of Model I and Model II are comparable to baseline on text chunking data and ATIS data, and worse than the baseline on LARGE data, while Model III gains higher segment-F1 score than baseline, demonstrating that the use of IOB labels is not suitable for building segmentation models independently. 
Moreover, Model I and II do not give consistent improvements on the final F1 score - the segmentation step improves labeling on slot filling, but not on the text chunking task.
Finally, Model III consistently performs better than baseline and gets state-of-the-art performance on the two tasks.
We also gain insights about the datasets we use by comparing the segment-F1 scores and F1 scores of model III.
For the text chunking data (95.75 vs. 94.72) and LARGE data (82.44 vs. 78.49),
the scores are close to each other, indicating that segmentation is a major challenge in these two datasets compared to labeling. 
But for ATIS data (99.01 vs. 95.86), the segmentation score is almost 100 percent,
so labeling seems like the main challenge in this dataset. 
We hope this insight encourages more research efforts on the similar tasks.
Finally, the proposed neural sequence chunking models achieves state-of-the-art performance on both text chunking and slot filling.

\bibliographystyle{aaai} 
\bibliography{refs}

\end{document}